# Coupling Adversarial Learning with Selective Voting Strategy for Distribution Alignment in Partial Domain Adaptation

Sandipan Choudhuri, Hemanth Venkateswara, Arunabha Sen
Arizona State University
{s.choudhuri,hemanthv,asen}@asu.edu

## ABSTRACT

In contrast to a standard closed-set domain adaptation task, partial domain adaptation setup caters to a realistic scenario by relaxing the identical label set assumption. The fact of source label set subsuming the target label set, however, introduces few additional obstacles as training on private source category samples thwart relevant knowledge transfer and mislead the classification process. To mitigate these issues, we devise a mechanism for strategic selection of highly-confident target samples essential for the estimation of class-importance weights. Furthermore, we capture class-discriminative and domain-invariant features by coupling the process of achieving compact and distinct class distributions with an adversarial objective. Experimental findings over numerous cross-domain classification tasks demonstrate the potential of the proposed technique to deliver superior and comparable accuracy over existing methods.

## 1 INTRODUCTION

The effectiveness of deep-classification networks is conditioned on the accessibility of annotated data which is, often, not readily available. Domain Adaptation (*DA*) methods [12, 7] mitigate this annotation demand by transferring knowledge from an already-labeled dataset in a related domain. A majority of the existing *DA* approaches [12, 7, 6] assumes that labeled and unlabelled domains share identical label set. In practice, however, obtaining a suitably labelled domain (source) with such a tight restriction on the label space is challenging. Partial domain adaptation (*pda*) [1] relaxes this constraint by utilizing a large-scale source domain data that subsumes the target label space.

Such a relaxation introduces the issue of superfluous information transfer from samples private to source domain (*negative transfer*)[1, 3], thereby thwarting the classification performance. Existing solutions [1, 20, 2, 3] have attempted to down-weight such samples by re-weighting them or by performing category-level aggregation of all target sample predictions, for estimating class-importance. However, poor classification at the initial stages of training may induce severe errors in such estimations, thereby misleading the optimization process. In this work, we address the negative transfer problem by strategically employing a subset of the target samples for adaptive estimation of class-importance weights.

Prior works [1, 20, 2, 3, 4] have focused on the effectiveness of aligning the latent distributions of source and target via an adversarial objective. Achieving domain invariance is, however, a necessary but not a sufficient condition for improving target classification performance; mitigating the conditional distribution mismatch across two domains is equally essential [10]. Citing this, we have designed a strategy to extract both *category-discriminative* and *domain-invariant features* by coupling an adversarial objective with a process aimed for obtaining more compact class distributions.

## 2 RELATED WORK

A series of works, in recent years, have explored the effectiveness of deep-neural architectures for mitigating domain shift and transferring underlying knowledge across domains in domain adaptation tasks [9, 14, 19]. Different line of works [12, 13] have utilized high-order statistical properties, like maximum-mean discrepancy, to align the data distribution across domains and successfully eliminate domain discrepancy. Authors in [7, 11] have utilized adversarial learning to design a mini-max game for obtaining domain-invariant features. Unfortunately, they operate on a restrictive *uda* environment and does not scale satisfactorily in a *pda* setup. To tackle *pda* tasks, Selective Adversarial Network (SAN) [1] utilizes multiple adversarial networks to down-weight private source category samples. The authors in [2] extend this concept to formulate class-importance weighting using all target samples. Importance Weighted Adversarial Nets (IWAN) [20] employ an auxiliary domain discriminator to determine the membership of a source sample in the target domain. Example Transfer Network (ETN) [3] operates on similar lines, by exploiting discriminative information as the transferability quantification for source domain samples. Albeit surpassing the performance of standard domain adaptation approaches, poor classification at the initial stages of training of these models may induce severe errors in private source category estimations, thereby misleading the classification task.

## 3 METHODOLOGY

### 3.1 Problem Settings

A typical *Unsupervised Domain-Adaptation* (uda) setup assumes that the source $s$ and target $t$ samples are drawn from different probability distributions [16]. As witnessed in such an environment, we are furnished with a source dataset $\mathcal{D}_s = \{\langle x_s^i, y_s^i \rangle\}_{i=1}^{|\mathcal{D}_s|}$ of $|\mathcal{D}_s|$ labeled points, sampled from a distribution $\mathbb{P}_s$, and an unlabeled dataset $\mathcal{D}_t = \{x_t^j\}_{j=1}^{|\mathcal{D}_t|}$ of $|\mathcal{D}_t|$ samples, representing target $t$, drawn from distribution $\mathbb{P}_t$ ($x_s^i, x_t^j \in \mathbb{R}^d$ and $\mathbb{P}_s \neq \mathbb{P}_t$). Since target class label information is unavailable during adaptation, the *Closed-Set* variant assumes samples in $\mathcal{D}_s$ and $\mathcal{D}_t$ categorized into classes from known label-sets $C_s$ and $C_t$ respectively, where $C_s = C_t$. *Partial DA* (pda) generalizes this characterization and addresses a realistic scenario by alleviating the constraint of identical label space assumptions between the two domains (i.e., $C_t \subseteq C_s$).

We aim to design a classifier hypothesis $h : x_t \rightarrow y_t$ ($h \in \mathcal{H}$) that minimizes the target classification risk under a *PDA* setup. This is achieved by leveraging source domain supervision to capture class-semantic information, along with minimizing misalignment due to negative transfer from samples $\langle x_s, y_s \rangle \in \mathcal{D}_{\bar{s}}$ in label-space $C_s - C_t$, where $\mathcal{D}_{\bar{s}} = \{\langle x_s, y_s \rangle | \langle x_s, y_s \rangle \in \mathcal{D}_s, y_s \in C_s - C_t\}$.

### 3.2 Proposed Approach

In this section, we present the proposed framework that couples adversarial learning with a selective voting strategy to mitigate domain discrepancy, while achieving class-distribution alignment.

*3.2.1 Domain-Invariant Feature Extraction with Adversarial Learning.* The idea of coupling adversarial learning with domain invariant feature extraction has been extensively researched in domain adaptation problems [7, 1]. Motivated by the effectiveness of DANN (Domain Adversarial Neural Network) [7] frameworks, we aim to achieve domain alignment by formulating a similar two-player mini-max game to match the source and target feature distributions. In the proposed setup, the first player $G_d(\cdot)$ is modelled as a domain classifier, and is trained to separate $x_s$ from $x_t$. The feature extractor, $F(\cdot)$, poses as the second player, and is designed to confuse the domain discriminator at the same time by generating domain-invariant features. The overall objective of the Domain Adversarial Neural Network is realized by minimizing the following term:

$$\mathcal{L}_{adv} = -\frac{1}{|\mathcal{D}_s|} \sum_{x_s \in \mathcal{D}_s} w_{y_s} \Big[ \log G_d(F(x_s)) \Big]$$
$$- \frac{1}{|\mathcal{D}_t|} \sum_{x_t \in \mathcal{D}_t} \Big[ 1 - \log G_d(F(x_t)) \Big] \quad (1)$$

The negative transfer, potentially caused by source samples in $\mathcal{D}_{\bar{s}}$ (section 3.1), is checked by regulating the contributions of all source samples. The underlying intuition centers around down-weighting the proportion of learning from private source samples. This is achieved by utilizing weights $w_{y_s}$ reflecting class-importance, estimated over the target samples (detailed process presented in section 3.2.3).

*3.2.2 Classification.* The feature extractor output is processed by a label classifier $G_y(\cdot)$, trained using the categorical cross-entropy loss $\mathcal{L}_{cl}(\cdot, \cdot)$ between the ground-truth labels of source samples and their predicted classification results. Similar to the domain adversarial training as presented above, the learning process is regulated by the class-importance weight (see section 3.2.3) estimated on target samples. The overall classification loss function is represented as:

$$\mathcal{L}_{class} = \frac{1}{|\mathcal{D}_s|} \sum_{x_s \in \mathcal{D}_s} w_{y_s} \mathcal{L}_{cl}\Big(G_y(F(x_s)), y_s\Big) \quad (4)$$

*3.2.3 Target Supervision and Estimating Class-Importance Weights through Pseudo-Labels.* For eliminating the negative influence of samples private to $s$, we propose a strategy of regulated learning where the model, adaptively, selects out a subset of the target domain samples that are share a high degree of similarity to the source domain. Inspired by the works in domain adaptation [12, 13], we enable target domain supervision by incorporating pseudo-labels, generated from a non-parametric classifier. The adopted pseudo-labelling strategy is as follows:

- **Step 1:** For an input sample $x_{s/t}$ in $\mathcal{D}_{s/t}$, we obtain its encoded latent representation $F(x_{s/t})$, using the feature extractor $F(\cdot)$.
- **Step 2:** With the encoded representations $F(x_s)$ of source entries $\langle x_s, y_s \rangle \in \mathcal{D}_s$, we compute the cluster centers $\{\mu_s^c\}$, $\forall c \in C_s$, where:

$$\mu_s^c = \frac{1}{|\mathcal{D}_s^c|} \sum_{x_s^c \in D_s^c} F(x_s^c) \quad (2)$$

Where $D_s^c = \{\langle x_s, y_s \rangle | \langle x_s, y_s \rangle \in \mathcal{D}_s, y_s = c\}$ ($|\cdot|$ represents set-cardinality).

- **Step 3:** A similarity function $\Phi(\cdot)$ (returns a vector of size $|C_s|$) processes each $F(x_t), x_t \in \mathcal{D}_t$, that quantifies it's closeness to the representative centers of each source class, and is represented as:

$$\Phi(F(x_t)) = \left\langle \frac{2 - \mathcal{JS}(F(x_t) || \mu_s^c)}{2} \right\rangle_{c \in C_s} \quad (3)$$

As highlighted above, we utilize Jensen-Shannon divergence, $\mathcal{JS}(\cdot || \cdot)$, to measure the divergence between the latent-target vector and representative centers of the source categories in latent space. Each divergence entry is normalized to [0,1], with a higher value signifying greater similarity.

- **Step 4:** Probability predictions for a target sample $x_t$ is computed using a $softmax(\cdot)$ over $\Phi(F(x_t))$. i.e.:

$$\mathbf{p_t} = softmax(\Phi(F(x_t))) \quad (4)$$
$$y_t^p \leftarrow \underset{c}{\operatorname{argmax}} \big[softmax(\Phi(F(x_t)))\big] \quad (5)$$

$y_t^p$, in eq. 5, represents the predicted pseudo-label for $x_t$, essential for optimizing the *inter* and *intra-class* discrepancies (refer to sections 3.2.4 and 3.2.5). $\mathbf{p_t}$, as highlighted in eq. 4, represents a vector of *softmax* probabilities, with $k^{th}$ entry representing the probability of $x_t$ belonging to the $k^{th}$ category in $|C_s|$ classes.

The confidence probability, $max(\mathbf{p_t})$ offers a monitoring mechanism of how likely the target sample $x_t$ is to be mapped with its nearest cluster center. A low value indicates a significant degree of confusion existing in the model, while classifying $x_t$ to a category in $C_s$. Using unreliable target samples with low confidence values for class-importance weight estimation might sabotage classification by diverting attention away from the desired objective. To eliminate such issues, we adopt a voting strategy for computing the class-importance weight vector $\mathbf{W}$ (with $|C_s|$ elements), in which only a subset of the target samples (those with high confidence predictions) can participate. The dataset $\mathcal{D}_\mathcal{T}$, thus constructed with these highly-confident target samples, is mathematically represented as:

$$\mathcal{D}_\mathcal{T} = \left\{ \langle x_t, y_t^p \rangle \mid x_t \in \mathcal{D}_t, y_t^p \leftarrow \operatorname*{argmax}_c(\mathbf{p_t}), max(\mathbf{p_t}) \geq \mathcal{T} \right\} \quad (6)$$

The threshold parameter, $\mathcal{T}$, is computed over the predicted outputs of the non-parametric classifier $softmax(\Phi(\cdot))$ and represents the average probability of source domain samples belonging to the ground-truth class, i.e.:

$$\mathcal{T} = \frac{1}{|\mathcal{D}_s|} \sum_{x_s \in \mathcal{D}_s} max\left(softmax\bigl(\Phi(F(x_s))\bigr)\right) \quad (7)$$

The class-importance weight vector $\mathbf{W} = [w_{c_1}, w_{c_2}, ..., w_{c_k}, ..., w_{c_{|C_s|}}]$ is estimated using the equation below:

$$\mathbf{W} \leftarrow \frac{\mathbf{W}'}{max(\mathbf{W}')}, \text{ where } \mathbf{W}' = \frac{1}{|\mathcal{D}_t^\mathcal{T}|} \sum_{x_t \in \mathcal{D}_t^\mathcal{T}} \mathbf{p_t} \quad (8)$$

In the proposed setup, given a source instance $\langle x_s, y_s \rangle \in \mathcal{D}_s$, where $y_s = c_k$, the corresponding class weight $w_{y_s=c_k}$ in $\mathbf{W}$ regulates the degree of learning from instance $\langle x_s, y_s \rangle$, as noted in sections 3.2.1 and 3.2.2. To ensure convenience in representation, the class weight $w_{y_s=c_k}$ is collapsed to $w_{y_s}$ in prior sections.

### 3.2.4 Maximizing inter-class separation.
Having a sufficiently "well-organized" and "regular" latent space arrangement is conducive to realizing the goal of achieving superior classification performance. Therefore, it is essential that samples with same class label are clustered to their respective distribution, while data with different class labels are allocated to distinct class distributions, regardless of their domains. We reinforce this concept by maximizing the $L_2$-distance between the mean embeddings of two distinct classes. It's worth noting that the approach also seeks to maximize the distance between different classes within the same domain (captured by the first two terms in eq. 9). This objective is partially realized using the *between-class loss* $\mathcal{L}_{bc}$, as presented below:

$$\mathcal{L}_{bc} = -\left[ \alpha \left( \frac{1}{|C_s|(|C_s|-1)} \sum_{c \in C_s} \sum_{\substack{c' \in C_s \\ c' \neq c}} \left\| \mu_s^c - \mu_s^{c'} \right\|_2 \right. \right.$$
$$+ \frac{1}{|C_\mathcal{T}|(|C_\mathcal{T}|-1)} \sum_{c \in C_\mathcal{T}} \sum_{\substack{c' \in C_\mathcal{T} \\ c' \neq c}} \left\| \mu_\mathcal{T}^c - \mu_\mathcal{T}^{c'} \right\|_2 \right)$$
$$\left. + \frac{\beta}{|C_\mathcal{T}|(|C_\mathcal{T}|-1)} \sum_{c \in C_\mathcal{T}} \sum_{\substack{c' \in C_\mathcal{T} \\ c' \neq c}} \left\| \mu_s^c - \mu_\mathcal{T}^{c'} \right\|_2 \right] \quad (9)$$

Here, $||\cdot||_2$ represents $L_2$-norm. In the equation above, $C_\mathcal{T}$ represents the label set of $D_\mathcal{T}$ (refer to eq. 6), where $C_\mathcal{T} \subseteq C_s$. We follow a similar routine as presented in eq. 2, for computing the mean-embedding $\mu_{s/\mathcal{T}}^c$ for a class $c$, over samples in dataset $D_{s/\mathcal{T}}$ respectively. Hyper-parameters $\alpha$ and $\beta$ control the contribution of cross-domain and within-domain components, in eq. 9.

### 3.2.5 Minimizing within-class separation.
As previously stated, in addition to maximizing the distance between two different clusters, it is essential to group samples of the same class together, to avoid classification errors on instances near the decision boundaries. We incorporate this concept into our framework by minimizing the distance between the embedded representations of any two samples in the same category, in a domain-agnostic setup. The *within-class loss* $\mathcal{L}_{wc}$, capturing this formulation, is presented as:

$$\mathcal{L}_{wc} = \frac{1}{|C_s|} \sum_{c \in C_s} \left[ \frac{1}{|\mathcal{D}^c|(|\mathcal{D}^c|-1)} \sum_{x^i \in \mathcal{D}^c} \sum_{\substack{x^j \in \mathcal{D}^c \\ x^j \neq x^i}} \left\| F(x^i) - F(x^j) \right\|^2 \right] \quad (10)$$

Where, dataset $\mathcal{D}^c = \{\langle x, y \rangle | \langle x, y \rangle \in \mathcal{D}, y = c \in C_s\}$, with $\mathcal{D} = \mathcal{D}_s \cup \mathcal{D}_\mathcal{T}$. In equation 10, we use indices $i, j$ to represent distinct samples in $\mathcal{D}^c$.

### 3.2.6 Entropy Minimization of Target Samples.
The early stages of a classification process in a domain adaptation setup witness two major adverse-effects: (a) difficulty in transferring sample information due to large domain shifts, and (b) inducing uncertainty in the classifier. To circumvent such issues, we adopt the entropy minimization principle during model training, which is formulated as:

$$\mathcal{L}_{em} = -\frac{1}{|\mathcal{D}_t|} \sum_{x_t \in \mathcal{D}_t} \sum_{c \in C_s} \bigl[G_y(F(x_t))\bigr]^c log\bigl(\bigl[G_y(F(x_t))\bigr]^c\bigr) \quad (11)$$

Where, $\bigl[G_y(F(x_t))\bigr]^c$ represents the probability of $x_t$ belonging to class $c$, as predicted by the classifier $G_y(\cdot)$.

### 3.2.7 Overall Objective.
To sum up, we propose the overall objective function as:

$$\mathcal{L} = \mathcal{L}_{class} + \eta \mathcal{L}_{adv} + \mathcal{L}_{bc} + \gamma \mathcal{L}_{wc} + \mathcal{L}_{em}, \quad (12)$$

with $\eta$ and $\gamma$ as user-defined hyper-parameters controlling the contribution of each loss in model learning.

## 4 EXPERIMENTS

In this section, we perform experiments on two benchmark datasets, (Office-Home[18] and Office-31[15]), to evaluate the efficacy of the proposed framework. The experiments are conducted in a *pda* setup, over multiple tasks for each dataset. Results of these tasks are reported and analyzed in the following sections.

### 4.1 Datasets

For overall performance assessment, we utilize standard datasets for domain adaptation, specifically *Office-Home* and *Office-31*. The Office-31 [15] dataset is grouped into 31 distinct categories, representing three domains: Amazon (A), DSLR (D) and Webcam (D). For evaluation task setup, we replicate the technique adopted by Cao et. al. [2], where the dataset representing the target domain contains images from 10 distinct classes. The assessment is conducted on the following source-target combinations: A→W, A→D, W→A, W→D, D→A and D→W. The larger Office-Home dataset [18] is grouped into 4 distinct domains: Artistic (Ar), Clip Art (Cl), Product (Pr) and Real-world (Rw). Following the PADA setup [2], we construct the source and target datasets from 65 and 25 different image classes respectively. 12 different permutations of source-target are used for evaluation purposes, namely Ar→Cl, Ar→Pr, Ar→Rw, Cl→Ar, Cl→Pr, Cl→Rw, Pr→Ar, Pr→Cl, Pr→Rw, Rw→Ar, Rw→Cl and Rw→Pr.

| Method | Ar → Cl | Ar → Pr | Ar → Rw | Cl → Ar | Cl → Pr | Cl → Rw | Pr → Ar | Pr → Cl | Pr → Rw | Rw → Ar | Rw → Cl | Rw → Pr | Avg. |
|---|---|---|---|---|---|---|---|---|---|---|---|---|---|
| Resnet-50[8] | 46.33 | 67.51 | 75.87 | 59.14 | 59.94 | 62.73 | 58.22 | 41.79 | 74.88 | 67.40 | 48.18 | 74.17 | 61.35 |
| DAN[12] | 43.76 | 67.90 | 77.47 | 63.73 | 58.99 | 67.59 | 56.84 | 37.07 | 76.37 | 69.15 | 44.30 | 77.48 | 61.72 |
| DANN[7] | 45.23 | 68.79 | 79.21 | 64.56 | 60.01 | 68.29 | 57.56 | 38.89 | 77.45 | 70.28 | 45.23 | 78.32 | 62.82 |
| ADDA[17] | 45.23 | 68.79 | 79.21 | 64.56 | 60.01 | 68.29 | 57.56 | 38.89 | 77.45 | 70.28 | 45.23 | 78.32 | 62.82 |
| PADA[2] | 51.95 | 67.00 | 78.74 | 52.16 | 53.78 | 59.03 | 52.61 | 43.22 | 78.79 | 73.73 | 56.60 | 77.09 | 62.06 |
| RTN[13] | 49.31 | 57.70 | 80.07 | 63.54 | 63.47 | 73.38 | 65.11 | 41.73 | 75.32 | 63.18 | 43.57 | 80.50 | 63.07 |
| IWAN[20] | 53.94 | 54.45 | 78.12 | 61.31 | 47.95 | 63.32 | 54.17 | 52.02 | 81.28 | 76.46 | 56.75 | 82.90 | 63.56 |
| SAN[1] | 44.42 | 68.68 | 74.60 | 67.49 | 64.99 | 77.80 | 59.78 | 44.72 | 80.07 | 72.18 | 50.21 | 78.66 | 65.30 |
| ETN[3] | 52.02 | 63.64 | 77.95 | 65.66 | 59.31 | 73.48 | 70.49 | 51.54 | 84.89 | 76.25 | 60.74 | 80.86 | 68.07 |
| Proposed model | **57.14** | **75.45** | **83.32** | 63.60 | **67.69** | 72.76 | 67.39 | **52.69** | 83.33 | 74.24 | 60.13 | **81.55** | **69.94** |
| W/o $\mathcal{L}_{adv}$ | 55.40 | 74.08 | 81.47 | 58.61 | 63.86 | 67.64 | 61.07 | 49.11 | 80.66 | 72.04 | 57.65 | 79.21 | 66.73 |
| W/o $\mathcal{L}_{bc}$ & $\mathcal{L}_{wc}$ | 53.25 | 70.53 | 77.76 | 56.24 | 60.87 | 64.59 | 58.48 | 47.21 | 77.92 | 69.16 | 55.47 | 76.83 | 64.02 |
| W/o $sv$ | 56.34 | 74.79 | 82.07 | 59.48 | 64.82 | 68.42 | 61.91 | 50.04 | 81.26 | 72.68 | 58.45 | 80.02 | 67.52 |

**Table 1: Classification accuracy (%) for Partial Domain Adaptation Tasks on Office-Home dataset (backbone: Resnet-50).**

| Method | A → W | A → D | W → A | W → D | D → A | D → W | Avg. |
|---|---|---|---|---|---|---|---|
| Resnet-50[8] | 75.59 | 83.44 | 84.97 | 98.09 | 83.92 | 96.27 | 87.05 |
| DAN[12] | 59.32 | 61.78 | 67.64 | 90.45 | 74.95 | 73.90 | 71.34 |
| DANN[7] | 73.56 | 81.53 | 86.12 | 98.73 | 82.78 | 96.27 | 86.50 |
| ADDA[17] | 75.67 | 83.41 | 84.25 | 99.85 | 83.62 | 95.38 | 87.03 |
| PADA[2] | 86.54 | 82.17 | 95.41 | **100.00** | 92.69 | **99.32** | 92.69 |
| RTN[13] | 78.98 | 77.07 | 89.46 | 85.35 | 89.25 | 93.22 | 85.56 |
| IWAN[20] | 89.15 | 90.45 | 94.26 | 99.36 | **95.62** | **99.32** | 94.69 |
| SAN[1] | 90.90 | 94.27 | 88.73 | 99.36 | 94.15 | **99.32** | 94.96 |
| ETN[3] | 91.52 | 90.87 | 94.36 | 98.94 | 90.61 | 92.88 | 93.20 |
| Proposed model | **92.45** | **92.79** | **97.23** | 100 | 95.29 | 98.61 | **96.06** |
| W/o $\mathcal{L}_{adv}$ | 88.13 | 86.21 | 94.33 | 98.57 | 92.99 | 98.12 | 93.05 |
| W/o $\mathcal{L}_{bc}$ & $\mathcal{L}_{wc}$ | 85.52 | 83.57 | 91.60 | 97.54 | 90.27 | 96.87 | 90.89 |
| W/o $sv$ | 86.26 | 84.35 | 93.37 | 97.91 | 91.02 | 97.49 | 91.73 |

**Table 2: Classification accuracy (%) for Partial Domain Adaptation Tasks on Office-31 dataset (backbone: Resnet-50).**

### 4.2 Implementation Details

The feature extractor $F(\cdot)$ uses ResNet-50 as backbone network [8], pre-trained on ImageNet dataset [5]. Following DANN [7], we introduce a bottleneck layer of length 256 before the fully connected layers. This layer generates latent representations, utilized by $softmax(\Phi(\cdot))$, $G_y(\cdot)$, $G_d(\cdot)$ and during optimizing *between-class* separation and *within-class* compactness. Similar to PADA [2], we fine-tune $F(\cdot)$, and train $G_y(\cdot)$ and $G_d(\cdot)$ layers from scratch (learning rates of new layers set to 10 times of $F(\cdot)$). We use mini-batch stochastic gradient descent (SGD) with momentum of 0.9 and follow a similar learning rate update strategy, as in DANN [7]. **W** is initialized to 1 for all classes and is updated using eq. 8 after every 150 epochs. The $\eta$ is progressively increased from 0 to 1, as in DANN [7]. Parameters $\alpha$, $\beta$ and $\gamma$ are set to 0.2, 0.9, 0.7, and 0.1, 0.9, 1.7 for Office-31 [15] and Office-home [18] datasets respectively.

### 4.3 Comparison Models

The model is evaluated against state-of-the-art models for closed-set and partial domain adaptation tasks, namely Resnet-50 [8], Deep Adaptation Network (DAN) [12], Domain Adversarial Neural Network (DANN) [7], Adversarial Discriminative Domain Adaptation (ADDA) network [17], Residual Transfer Networks (RTN) [13], Importance Weighted Adversarial Nets (IWAN) [20], Selective Adversarial Network (SAN) [1], Partial Adversarial Domain Adaptation(PADA) [2] and Example Transfer Network (ETN) [3].

## 5 RESULTS AND ANALYSIS

In the results summarized in Tables I and II, the proposed method achieves comparable accuracy results to the state-of-the-art models addressing closed-set and partial domain adaptation, on all the presented tasks (achieving highest accuracy in 6 out of 12 tasks and in 4 out of 6 tasks on Office-Home and Office-31 datasets, respectively). It outperforms the existing methods in the overall *average accuracy percentage* on both datasets. For further analysis, we conducted an ablation study on the proposed model by suppressing its three main components, one at a time:

- **Without $\mathcal{L}_{adv}$**: We restrict the learning of domain-invariant latent representations by masking the adversarial objective (setting $\eta$ to 0).
- **Without $\mathcal{L}_{bc}$ and $\mathcal{L}_{wc}$**: The model is trained without class distribution alignment losses (setting $\alpha$, $\beta$, $\gamma$ to 0).
- **Without selective voting** ($sv$): We limit the utilization of highly confident target samples for computation of class-importance weights (setting threshold $\mathcal{T}$ to 0).

The reduction in classification performance, as witnessed in tables 1 and 2, after excluding these modules validates their significance in the classification task. Their utility is buttressed further with the model yielding best performance scores on a majority of the adaptation tasks described in section 4.1.

## 6 CONCLUSION

This paper presents a novel framework for partial domain adaptation tasks, that aims to mitigate domain discrepancy, while achieving class-distribution alignment in the latent space. The proposed approach couples adversarial learning with a selective consensus strategy (using a non-parametric classifier) that prevents a set of target samples, with low confidence predictions, from participating in class-importance weight estimation and alignment of category-level distributions. From the experiments conducted over two benchmark datasets, it is established that our approach achieves superior classification results, when compared to state-of-the-art models.